%% file: main.tex
\documentclass[10pt,twocolumn,letterpaper]{article}
\usepackage[pagenumbers]{cvpr}
\usepackage{graphicx}
\usepackage{amsmath}
\usepackage{amssymb}
\usepackage{booktabs}
\usepackage[inline]{enumitem}
\usepackage{tabularx}
\usepackage{multirow}
\usepackage[pagebackref,breaklinks,colorlinks]{hyperref}
\usepackage{ragged2e}
\usepackage[capitalize]{cleveref}
\crefname{section}{Sec.}{Secs.}
\Crefname{section}{Section}{Sections}
\Crefname{table}{Table}{Tables}
\crefname{table}{Tab.}{Tabs.}
\graphicspath{ {./Figures/} }
\input{settings}
\usepackage[hang,flushmargin]{footmisc} 
\makeatletter
  \renewcommand{\paragraph}{%
    \@startsection{paragraph}{4}%
    {\z@}{1.0ex \@plus 1ex \@minus .2ex}{-0.5em}
    {\normalfont\normalsize\bfseries}%
  }
  \def\NAT@spacechar{}
  \newcommand*{\toccontents}{\@starttoc{toc}}
  \pretocmd{\section}{\addtocontents{toc}{\vspace{-9pt}}}{}{}
  \DeclareRobustCommand{\sm@ller}{%
    \dimen@\f@size\p@
    \ifdim \dimen@ > 12\p@
    4
    \dimen@=0.83333\dimen@
    \else
    \advance \dimen@ -2\p@
    \fi
    \math@fontsfalse
    \fontsize{\the\dimen@}\z@
    \selectfont
  }
\makeatother

\everypar=\expandafter{\the\everypar\loosness=-1} 

\begin{document}

\title{\textit{EBMs vs. CL}: Exploring Self-Supervised Visual Pretraining for\\Visual Question Answering}

\author{Violetta Shevchenko$^1$, Ehsan Abbasnejad$^1$, Anthony Dick$^1$, Anton van den Hengel$^{1,2}$, Damien Teney$^{1,3}$ \\
$^1$University of Adelaide \hspace{5pt} $^2$Amazon \hspace{5pt} $^3$Idiap Research Institute\\
{\tt\small [firstname].[lastname]@adelaide.edu.au}
}

\maketitle

\begin{abstract}
\noindent
\textbf{Context.}
The availability of clean and diverse labeled data is a major roadblock for training models on complex tasks such as visual question answering (VQA).
The extensive work on large vision-and-language models has shown that self-supervised learning is effective for pretraining multimodal interactions.
In this technical report, we focus on visual representations.
We review and evaluate self-supervised methods to leverage unlabeled images and pretrain a model, which we then fine-tune on a custom VQA task that allows controlled evaluation and diagnosis.
We compare energy-based models (EBMs) with contrastive learning (CL).
While EBMs are growing in popularity, they lack an evaluation on downstream tasks.


\noindent
\textbf{Findings.}
Both EBMs and CL can learn representations from unlabeled images that enable training a VQA model on very little annotated data.
In a simple setting similar to CLEVR, we find that CL representations also improve systematic generalization, and even match the performance of representations from a larger, supervised, ImageNet-pretrained model.
However, we find EBMs to be difficult to train because of instabilities and high variability in their results.
We, therefore, investigate other purported benefits of EBMs.
They prove useful for OOD detection, but other results on supervised energy-based training and uncertainty calibration are largely negative.

\noindent
\textbf{Conclusions.}
(1)~We make the encouraging observation that self-supervised visual pretraining allows displacing some of the requirements for training data from the main/supervised to a pretraining/unsupervised stage. (2)~CL currently seems a preferable option over EBMs.
To our surprise, EBMs could not achieve the benefits purported in the literature, even in a toy setting.
\end{abstract}

\section{Introduction}
\label{sec:intro}

The availability of large-scale diverse datasets has been a driving force in deep learning research for computer vision as well as natural language processing (NLP). Tasks like visual question answering (VQA), that combine vision and language and involve complex reasoning, require a large amount of aligned visual and textual data for training.
The generation of such data involves human annotators, making it expensive and time-consuming. Moreover, task-specific data collection from humans often introduces biases and noise in the data~\cite{kafle2017visual,kafle2019challenges} that are subsequently difficult to deal with during training~\cite{teney2020learning,teney2020unshuffling} or evaluation~\cite{dancette2021beyond,teney2020value}.

Early methods for VQA made use of pretrained image models~\cite{he2016deep,anderson2018bottomup} and language features~\cite{pennington2014glove,mikolov2013efficient}.
Most recent work has focused on pretraining general-purpose vision-and-language models~\cite{tan2019lxmert,li2019visualbert,li2020oscar,chen2019uniter}.
These models are pretrained with self-supervised objectives, which alleviates the need for tasks-specific VQA training examples, but they usually require aligned vision and language, such as images with captions.
This motivates our exploration of self-supervised methods for pretraining the visual encoder, which only requires unlabeled images.

Our study explores the potential benefits of two popular self-supervised paradigms, energy-based models (EBMs) and contrastive learning (CL).
These are applied specifically to \emph{visual} data.
Our findings are therefore particularly relevant to the training of vision-and-language models for particular visual domains (\eg industrial or medical images) where little annotated data is available.
We conduct our experiments with a simple VQA model and a custom toy task similar to CLEVR~\cite{johnson2017clevr}, which allows a controlled evaluation.
Unlike most recent papers on vision and language, we do not deal with large-scale datasets nor large transformer-based models.

An important novelty of this study is the consideration of EBMs.
Generative models are generally attractive in unsupervised machine learning~\cite{kingma2013auto,goodfellow2014generative}.
 EBMs in particular have grown in popularity~\cite{grathwohl2019your,du2019implicit,zhao2020joint}
because they were shown to be also effective on discriminative tasks, with claims of improved calibration, robustness, and out-of-distribution (OOD) detection capabilities.
The current literature however lacks evaluations on complex downstream tasks, which we partly remedy.

\vspace{5pt}
\noindent
The contents of this report are summarized as follows.
\setlist{nolistsep,leftmargin=*}
\begin{enumerate}[topsep=-2pt,itemsep=5pt]
  \item We review the general concepts behind energy-based models (EBMs) and contrastive learning (CL).
  \item We devise a simple VQA setting with data similar to CLEVR, and a simple CNN--LSTM model. Both allow a constrained and controlled evaluation, including OOD generalization (novel object/color combinations at test time).

  \item In this setting, we compare EBMs with CL (SimCLR~\cite{chen2020simple}) for pretraining visual representations used by the VQA model. Both methods are effective and allow training the VQA model with little annotated data.
  However, EBMs are practically difficult to train.

  \item We compare EBMs with CL for OOD detection of images, which could serve in VQA for determining unanswerable cases at test time.
  EBMs are generally more reliable, but they can fail when ID and OOD data are visually similar.

  \item We investigate potential benefits of supervised energy-based training of the VQA model itself, using JEM~\cite{grathwohl2019your} and CEBM~\cite{du2020improved}.
  Our results are mixed, with some improvements in accuracy (at the expense of a finicky training), but none of the expected improvements in calibration.
\end{enumerate}

\section{Related work}
\label{sec:related}

\paragraph{Training VQA models.}
Visual question answering is a popular task to evaluate the progress of AI on vision and language~\cite{teney2017visuala,wu2016survey}.
Early work on VQA trained models with supervision from question/image/answer triplets. These models (e.g.~\cite{teney2017tips}) used visual features from a CNN or object detector~\cite{anderson2018bottomup} itself trained with supervision from annotated datasets such as ImageNet~\cite{deng2009imagenet} or Visual Genome~\cite{krishna2017visual}.
Modern approaches to VQA use large-scale transformer models trained with self-supervised objectives on vision-and-language data~(e.g.~\cite{tan2019lxmert}), before being fine-tuned with supervision of VQA triplets.
The self-supervision in these models serves to learn correspondences across the visual and textual modalities, and the visual inputs are often processed by a pretrained visual encoder. Recent exceptions that appeared after our study include ViLT~\cite{kim2021vilt} and VLC~\cite{gui2022training}, which take image patches as input, and
MDETR~\cite{kamath2021mdetr}, X-DETR~\cite{cai2022x}, SOHO~\cite{huang2021seeing}, TxT~\cite{steitz2021txt} and E2E-VLP~\cite{xu2021e2e}, which jointly train an object detector.

\paragraph{Energy-based models}%
\label{subsec:related_ebm}%
(EBMs) offer an attractive framework to unify various areas of machine learning, and also lead to a new class of generative models.
See~\cite{lecun2006tutorial} for an early tutorial.
Recent studies~\cite{nijkamp2019learning,du2019implicit,nijkamp2020anatomy} have investigated training techniques that enable the application of EBMs to high-dimensional data and address issues of training stability.
EBMs have been applied to various areas including image generation~\cite{xie2018cooperative,du2019implicit,han2019divergence,du2020compositional,xiao2020vaebm,arbel2020generalized},
graph generation~\cite{suhail2021energy}, image classification~\cite{grathwohl2019your}, regression~\cite{gustafsson2020energy,gustafsson2020train}, continual learning~\cite{li2020energy} and natural language processing~\cite{deng2019residual,tu2020engine,he2021joint}.

\paragraph{Contrastive learning}%
\label{subsec:related_contr}%
(CL) is one of the most popular approaches to learn representations of high-dimensional data such as images without access to training labels.
A contrastive training objective essentially trains a model to differentiate between similar and dissimilar samples.
While early application in computer vision~\cite{wu2018unsupervised,oord2018representation,zhuang2019local,bachman2019learning,henaff2020data,tian2020contrastive,he2020momentum} could not compete with supervised training, recent methods such as SimCLR~\cite{chen2020simple} and SwAV~\cite{caron2020unsupervised} introduced novel data augmentations and architectural modifications that narrowed this gap and can now compete with the supervised paradigm.
In this work, we apply contrastive learning to the visual question answering task. Another VQA method that incorporates contrastive loss was proposed by Whitehead \etal~\cite{whitehead2021separating}, where the model is trained on image-question pairs in a self-supervised manner. In contrast, in this work we utilize images without any additional annotations.

\section{Background}
\label{sec:method}

The setup of our main experiments consists in pretraining a convolutional neural network (CNN) on unlabeled images with self-supervision.
This CNN is then used as the visual encoder of a VQA model.
The CNN is fine-tuned while training the VQA model with question/image/answer examples.
With this setup, we investigate whether visual pretraining can facilitate the training of the VQA model (\eg with less task-specific data) and/or improve its generalization capabilities.
Our work does not introduce new methods besides the adaptation of EBMs and contrastive learning to the pretraining of visual representations for VQA.

This section reviews the methods used under a common technical umbrella.
This background justifies the comparison of these methods in a same study and it sets the stage in terms of expectations for our experiments of \sectref{sec:experiments}.

\input{background}

\subsection{VQA Model}
\label{subsec:method_finetune}
In our main experiments we use a simple VQA model that is built from a pretrained visual encoder as follows.
We combine a visual encoder $f^I$ (a CNN) with a question encoder $f^Q$ (an LSTM).
They respectively process an image $\bx^I$ and question $\bx^Q$ into representations $\bv \in \mathbb{R}^\cardinality{D_I}$ and $\bq \in \mathbb{R}^\cardinality{D_Q}$:
\abovedisplayskip=5pt
\belowdisplayskip=5pt
\begin{equation}
\label{eq:method_finetune_encode}
    \bv = f_{\btheta}^I(\bx^I) ~~~~~~\textrm{and}~~~~~~ \bq = f_{\btheta}^Q(\bx^Q)
\end{equation}
These are concatenated into a single vector $\bp = [\bv, \bq]$, $\bp \in \mathbb{R}^\cardinality{D_I + D_Q}$. This vector is passed through an MLP classifier to obtain answer scores $\by=f^{CLS}(\bp)$, with $\by \in \mathbb{R}^\cardinality{A}$ where $A$ is the size of a set of candidate answers.
The model is trained with supervision on question/image/answer triplets to minimize a cross-entropy loss:
\abovedisplayskip=2pt
\belowdisplayskip=0pt
\begin{equation}
\label{eq:method_finetune_softmax}
    \softmax(y_{i}) ~=~ \frac{\exp(y_{i})}{\sum_{j=1}^{A}\exp(y_{j})} ~,
\end{equation}\vspace{-6pt}
\abovedisplayskip=4pt
\belowdisplayskip=0pt
\begin{equation}
\label{eq:method_finetune_loss}
    \mathcal{L}_{CE} ~=~ -\sum_{i=1}^{A} \hat{a}_{i} \cdot \log(\softmax(y_{i})) ~,
\end{equation}\vspace{-2pt}
where $\hat{\ba} \in \{0,1\}^\cardinality{A}$ denotes the one-hot (multi-hot) vector of the ground-truth answer(s), and $i$ indexes vector elements.

\section{Our implementation}
\label{subsec:experiments_setting}

\paragraph{Our VQA model} is a simple CNN--LSTM architecture, commonly used as a baseline in early work on VQA~\cite{antol2015vqa,malinowski2015ask,ren2015exploring}.
We settled on this simple option because it reduces the training instabilities arising when training EBMs within more complex architectures~\cite{du2019implicit, du2020improved}.
The CNN in this architecture is the EBM version of a ResNet~\cite{he2016deep} from Du \etal~\cite{du2020improved}.
The same ResNet backbone is used in all pretraining and fine-tuning experiments to enable transfer learning. We only modify its last few layers for each pretraining task, as discussed below.

\paragraph{For EBM pretraining,} we append the CNN model with a linear layer producing a scalar energy value. To facilitate the training we incorporate three changes proposed by Du \etal~\cite{du2020improved}. (1)~We use a replay buffer to store previously generated samples that are randomly chosen to reinitialize the sampling chain (\eqnref{eq:method_ebm_mcmc}) instead of the uniform noise. (2)~We apply random data augmentations (cropping, horizontal flipping, blurring, and color distortions) to images sampled from the buffer, which improves the diversity and mixing of the sampling chain. (3)~We include additional losses that improve the contrastive-divergence training (see~\cite{du2020improved} for details).

\paragraph{For contrastive pretraining,} we use the popular SimCLR method~\cite{chen2020simple}.
The standard SimCLR uses a standard ResNet-50 to extract visual representations and an additional projection head into the space where the contrastive loss is applied.
We swap the ResNet-50 for the one mentioned above and keep the standard projection head (a 1-layer MLP with ReLU activations).
This model is referred to as SimCLR$\star$ to emphasize the modified architecture. We use the same data augmentations as for EBM pretraining.
\section{Experiments}
\label{sec:experiments}

\input{Sections/tab_dataset}

\paragraph{Diagnostic dataset.}
\label{subsec:experiments_dataset}
Our main experiments use a custom controlled small-scale dataset, most similar to the existing CLEVR-CoGenT~\cite{johnson2017clevr}.
It similarly allows evaluating systematic generalization to unseen combinations of objects/colors.
The reason for a new small-scale dataset is our desire to evaluate EBMs, which remain computationally expensive at the time of this study (mid~2021).
Contemporary work on EBMs~\cite{du2020compositional,grathwohl2019your} indeed focuses on small datasets like MNIST~\cite{deng2012mnist}, CIFAR~\cite{krizhevsky2009learning} and CelebA~\cite{liu2015faceattributes}.

\input{Sections/fig_dataset}

We generate our dataset following the procedure of~\cite{atzmon2020causal}.
We render $64\!\times\!64$ images with Blender~\cite{blender2018}.
Each image contains one of three objects (sphere, cube, or cylinder) in one of eight colors (red, purple, yellow, blue, green, cyan, gray, or brown).
For each object/color combination, we generate 2000 samples, randomizing the object size, material (rubber or metallic), position, and lightning.
We assign disjoint subsets of object/color combinations to the pretraining, fine-tuning, and evaluation sets, as summarized in \tabref{tab:experiments_dataset}.
The validation set contains a mix of in-domain (ID) and out-of-distribution (OOD) data, while the test set if fully OOD. The validation set serves to select hyperparameters and perform early stopping.

For each image, we use templates from CLEVR to generate a question that queries the object shape (\eg \textit{There is a blue object in the image; what is it?}, with the three possible answers \textit{sphere}, \textit{cube} and \textit{cylinder}).
Because the range of questions is limited, \textbf{the input questions are in fact not necessary for a VQA model to find the correct answer. However they act as \emph{distractors} since they are spuriously correlated with the correct answers} because of mentioning colors.
A model relying on these spurious correlations would however perform badly on our test set because of the novel object/color combinations.
See \figref{fig:experiments_data} for examples.

\subsection{Generalization without/with pretraining}
\label{subsec:experiments_results}

\input{Sections/tab_accuracy}

We report our main results in \tabref{tab:experiments_acc}.
We compare models built with an EBM-- or CL--pretrained vision encoder, with a baseline that is not pretrained, \ie only trained with supervision on the VQA data.
The baseline overfits to the particular object/color combinations of the VQA data and is therefore unable to generalize to the novel combinations of the validation (ID+OOD data) nor test splits (fully OOD data).
Of our pretrained models, \textbf{the CL method (SimCLR$\star$) achieves the highest, near-perfect accuracy of 99.26\% on the test set.}
It seems to better disentangle the representations of shape from color, such that the VQA model can generalize to new combinations at test time without overfitting to the particular combinations seen during training.
EBM pretraining also improves generalization (86.74\% on the test set) but with a high variance across runs (standard deviation of 8.34\%).
SimCLR$\star$ shows no such stability issues (standard deviation of 0.29\%).

We also evaluate a model with ImageNet-pretrained ResNet-18 as the visual encoder, which is the most common paradigm for pretraining via transfer learning.
However the comparison is relatively unfair to our self-supervised models since this model relies on vastly more pretraining data as well as the ImageNet labels.
Still, SimCLR$\star$ performs on par with this model.

We also experimented with using our pretrained visual encoders frozen within a VQA model.
These results were always worse than with fine-tuning and are not included in the tables.

\subsection{Amount of pretraining data}

\begin{figure}[h!]
    \centering
    \begin{FlushRight}
    \includegraphics[height=95pt]{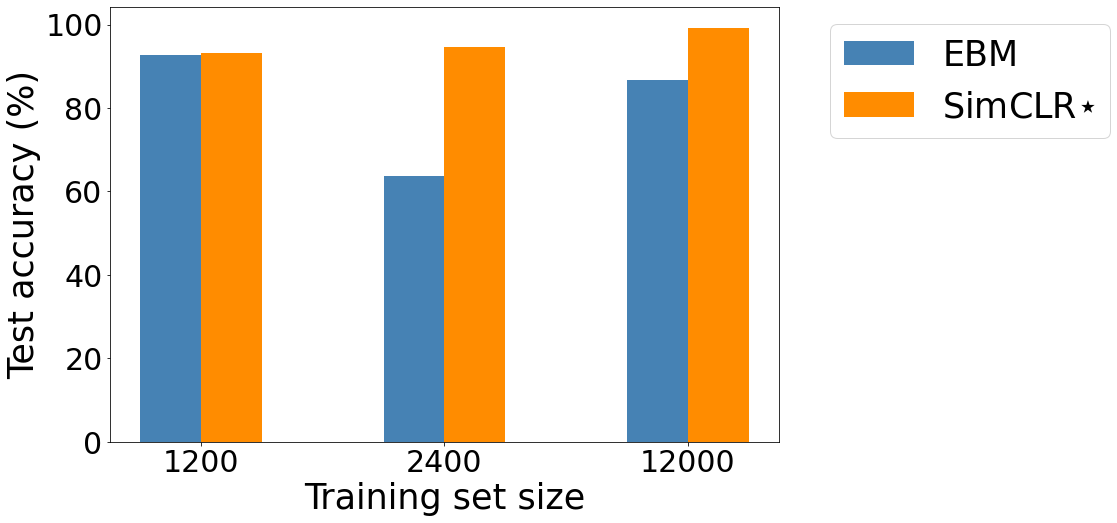}
    \end{FlushRight}\vspace{-10pt}\vspace{-5pt}
    \caption{\textbf{Decreasing the amount of pretraining data.}
The performance of SimCLR$\star$ model degrades only slightly even down to a 10-fold decrease.
SimCLR$\star$ also remains superior to EBM in all regimes.
The training stability of EBM proves again to be a challenge.
The high variability across runs even shows up as a slight improvement in performance with less pretraining data, although this is an artifact of these instabilities.}
    \label{fig:experiments_data_size_train}
\end{figure}

\begin{figure}[h!]
    \begin{FlushRight}
    \includegraphics[height=95pt]{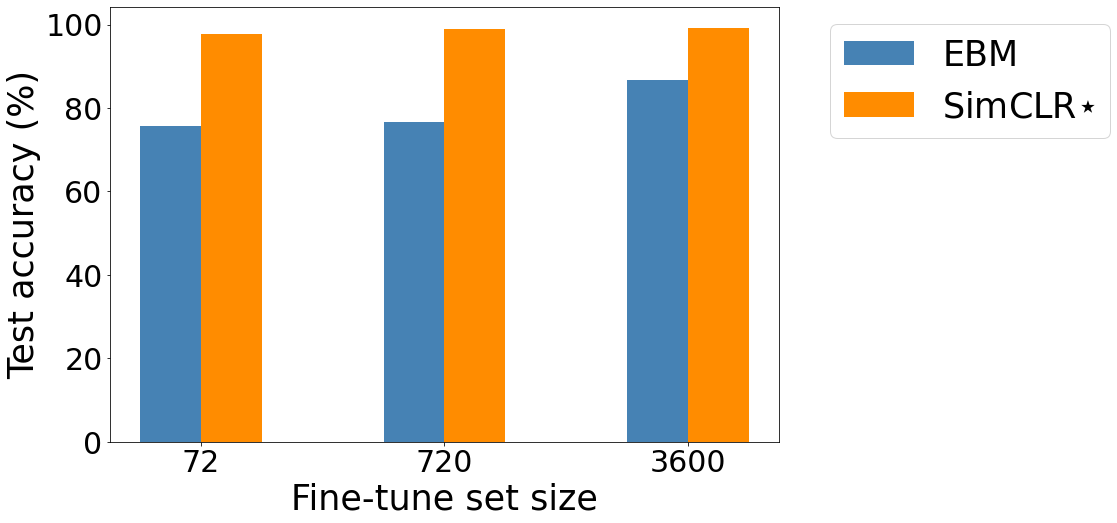}
    \end{FlushRight}\vspace{-10pt}\vspace{-5pt}
    \caption{\textbf{Decreasing the amount of labeled data.}
Both CL and EBM pretraining enable training the VQA model with very little data, even as few as 72 examples.
The accuracy of EBM drops only slightly, and that of SimCLR$\star$ model remains almost identical.}
    \label{fig:experiments_data_size_finetune}
\end{figure}

To better understand the benefits of pretraining in low-data regimes, we repeat our experiments while decreasing the amount of pretraining data (unlabeled images) or labeled training data (VQA examples).
See Figures~\ref{fig:experiments_data_size_train}--\ref{fig:experiments_data_size_finetune}.

\subsection{Out-of-distribution detection}
\label{subsec:experiments_ood}

OOD Detection is relevant to VQA because it can help determine whether a given image/question is answerable~\cite{chiu2020assessing,bhattacharya2019does}.
An image of low quality or from a different domain than the training examples (\eg clip art vs. photographs) can be detected as OOD and warn the user that the VQA model may not be reliable on this particular input.
A common baseline for OOD detection with a discriminative model (such as our VQA model pretrained with CL) is to compare the top softmax score with a predefined threshold. A low score means a low confidence hence an OOD sample~\cite{hendrycks2016baseline}.
EBMs have been purported to be useful for OOD detection~\cite{liu2020energy,elflein2021out} by their generative nature since they specifically model the distribution of the training data.
To use EBMs for OOD detection, one can simply compare the negative energy estimated by the model with a threshold.

\paragraph{Experimental setup.}
We evaluate our EBM and CL visual encoders from \sectref{subsec:experiments_results}.
As ID test data, we use the images from the test set of our toy VQA task.
As OOD data, we use images from various existing datasets (CIFAR~\cite{krizhevsky2009learning}, MNIST~\cite{deng2012mnist}, SVHN~\cite{netzer2011reading}, and VQA v2~\cite{goyal2017making}) as well as images of random noise, of cone shapes (of similar style as the toy VQA data), and of empty backgrounds (see \figref{fig:experiments_ood_ex}).
We compare our SimCLR$\star$ and EBM models using their softmax score and negative energy, respectively, for OOD detection. 
To quantify detection performance across possible thresholds, we compute the area under the ROC curve~(AUROC).

\begin{figure}[th]
    \centering
	\begin{subfigure}[t]{0.43\linewidth}
		\includegraphics[width=1\linewidth]{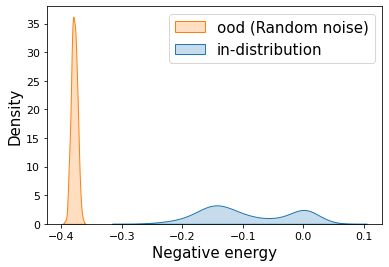}
		\caption{EBM (random noise)}
	\end{subfigure}
	\hspace{10pt}
	\begin{subfigure}[t]{0.43\linewidth}
		\includegraphics[width=1\linewidth]{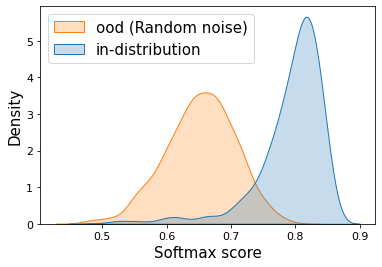}
		\caption{SimCLR$\star$ (random noise)}
	\end{subfigure} \\ \vspace{5pt}
	\begin{subfigure}[t]{0.43\linewidth}
		\includegraphics[width=1\linewidth]{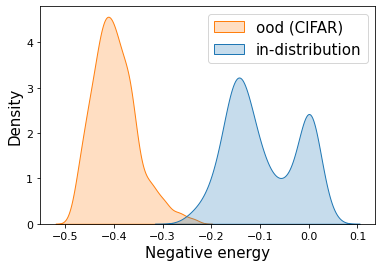}
		\caption{EBM (CIFAR)}
	\end{subfigure}
	\hspace{10pt}
	\begin{subfigure}[t]{0.43\linewidth}
		\includegraphics[width=1\linewidth]{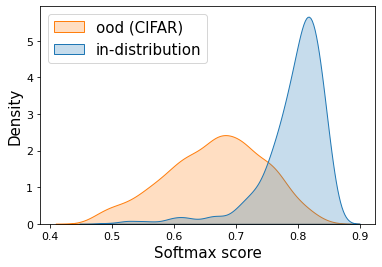}
		\caption{SimCLR$\star$ (CIFAR)}
	\end{subfigure} \\ \vspace{5pt}
	\begin{subfigure}[t]{0.43\linewidth}
		\includegraphics[width=1\linewidth]{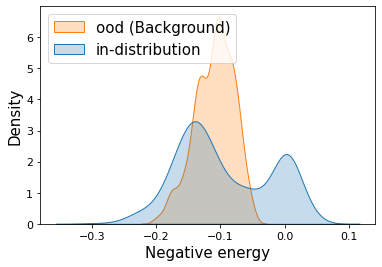}
		\caption{EBM (backgrounds)}
	\end{subfigure}
	\hspace{10pt}
	\begin{subfigure}[t]{0.43\linewidth}
		\includegraphics[width=1\linewidth]{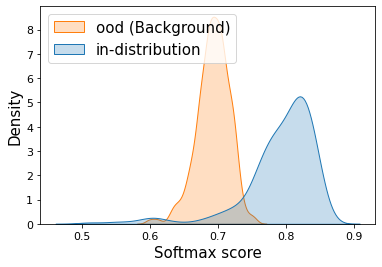}
		\caption{SimCLR$\star$ (backgrounds)}
	\end{subfigure} \\ \vspace{5pt}
	\begin{subfigure}[t]{0.43\linewidth}
		\includegraphics[width=1\linewidth]{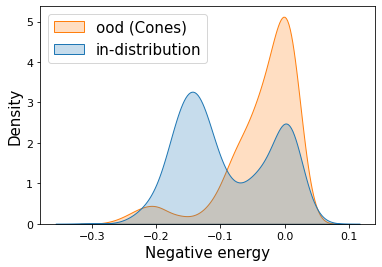}
		\caption{EBM (cones)}
	\end{subfigure}
	\hspace{10pt}
	\begin{subfigure}[t]{0.43\linewidth}
		\includegraphics[width=1\linewidth]{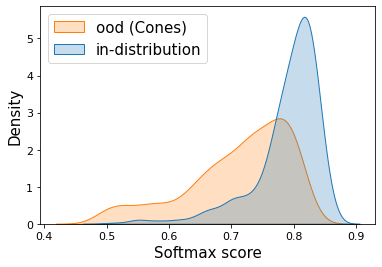}
		\caption{SimCLR$\star$ (cones)}
	\end{subfigure}
    \caption{OOD Detection experiment. Distributions of energy and softmax scores, respectively from an EBM~(left column) and a SimCLR$\star$ model~(right column) with test data from various sources~(rows).
    EBMs allow separating ID from OOD data almost perfectly by simple thresholding on some datasets (a,c).
    However they do poorly when ID and OOD data are visual similar (e,g).}
    \label{fig:experiments_ood}
\end{figure}

\paragraph{Results.}
On five out of our seven datasets, EBMs are superior to CL (see \tabref{tab:experiments_ood}).
They can distinguish ID from OOD data almost perfectly (AUROC of $\sim$1.0).
However, when the OOD data is visually similar to the training data (\eg images of backgrounds and cones), the detection performance of our EBM falls well below CL.
\textbf{The estimated energies appear generally reliable for filtering out OOD samples, but they seem less able to spot fine-grained differences.}
See also \figref{fig:experiments_ood} for the distributions of scores over different datasets.

\input{Sections/fig_ood_ex}

\begin{table*}[t]
    \small
    \centering
    \renewcommand\tabcolsep{4pt}
    \renewcommand{\arraystretch}{1}
    	\begin{tabularx}{0.8\textwidth}{Xccccccc}
          \toprule
          & \multicolumn{7}{c}{Source of OOD test data}\\
          \cmidrule{2-8}
          & Noise & CIFAR & MNIST & SVHN & VQA v2 & Background & Cones\\
          \midrule
          EBM & $\sim$\textbf{1.00} & $\sim$\textbf{1.00} & $\sim$\textbf{1.00} & $\sim$\textbf{1.00} & $\sim$\textbf{1.00} & 0.52 & 0.30 \\
          SimCLR$\star$ & ~~~0.96 & ~~~0.92 & ~~~0.97 & ~~~0.91 & ~~~0.94 & \textbf{0.94} & \textbf{0.82}\\
          \bottomrule
        \end{tabularx}
        \caption{OOD Detection experiment. The AUROC is near-perfect ($\sim$1.0) for our EBM model on most test sets.
        However, it falls well below our CL model when the OOD test data is visually similar to the training (ID) data (last two columns).}
    \label{tab:experiments_ood}
\end{table*}

\subsection{Supervised energy-based training}
\label{subsec:experiments_additional}

\begin{table*}[t]
    \small
    \centering
    \renewcommand\tabcolsep{2pt}
    \renewcommand{\arraystretch}{1}
    \begin{tabular}{l@{\hskip10pt} cccccc}
        \toprule
        ~ & \multicolumn{4}{c}{Answer accuracy (\%)} && Calibration\\
        \cmidrule{2-5} \cmidrule{7-7}
        ~ & Overall & Spheres & Cubes & Cylinders && ECE (\%) \\
        ~ & ~ & \scriptsize{(Seen combinations)} & \multicolumn{2}{c}{\scriptsize{(OOD object/color combinations)}} && \scriptsize{(Lower is better)} \\
        \midrule
        Random guessing & 33.34 & 33.34 & 33.34 & 33.34 && --- \\
        Baseline, cross-entropy classification objective & 33.34 & \textbf{100.00} & 0.00 & 0.00 && 71.54 \\
        \midrule
        Standard JEM~\cite{grathwohl2019your} & 29.98 & 99.93 & 0.00 & 0.00 && 71.28 \\
        Modified JEM: downscaled classification loss \& early stopping & \textbf{49.10} & 99.93 & \textbf{39.23} & 8.12 && 36.55 \\
        CEBM~\cite{du2020improved} & 35.75 & 54.60 & 31.56 & \textbf{21.10} && \textbf{9.24} \\
        \bottomrule
    \end{tabular}
    \caption{Supervised energy-based training of a VQA model. Modified JEM correctly classifies some of the OOD samples and improves calibration over the standard classification baseline. However, its performance is still far below optimal.}
    \label{tab:experiments_ebm_res}
\end{table*}

This section looks at energy-based methods for training the VQA model itself \ie not solely the visual encoder.
In our experiments so far, EBMs have proved generally inferior to CL.
This section investigates whether other purported capabilities of EBMs have value for VQA.
We look in particular at (1)~their ability to simultaneously address generative and discriminative tasks and (2)~their improved calibration compared to standard discriminative models.
We test two existing energy-based methods to train the VQA model with supervision from standard question/image/answer triplets.

\paragraph{Setup.}
We use a dataset similar to \sectref{subsec:experiments_dataset} that allows evaluating ID and OOD accuracy.
Every image/question contains either a sphere, a cube, or a cylinder.
Spheres can be of any color (ID test data),
whereas the sets of possible colors of cubes and cylinders are disjoint and swapped between training and test (OOD data).
Our baseline is the same CNN--LSTM as mentioned in \sectref{subsec:experiments_setting}.
The JEM and CEBM methods build on this same architecture.
We briefly review these methods before presenting our results.

\paragraph{The JEM approach}~\cite{grathwohl2019your} treats a discriminative classifier as an energy-based model by optimizing a double objective:
\abovedisplayskip=3pt
\belowdisplayskip=3pt
\begin{equation}
\label{eq:experiments_jem}
    \log p_{\btheta}(\bx, y) ~=~ \log p_{\btheta}(\bx) + \log p_{\btheta}(y | \bx)~,
\end{equation}
where $\bx$ is a training point and $y$ is its class label. This objective can be optimized with standard cross-entropy for classification part and log-likelihood (\eqnref{eq:method_ebm_log}) for energy learning. The energy function is defined as the negative $\text{LogSumExp}(\cdot)$ function of the logits of the classifier:
\abovedisplayskip=5pt
\belowdisplayskip=3pt
\begin{equation}
\label{eq:experiments_jem_energy}
    E_{\btheta}(\bx) ~=~ -\log \sum_{y} \exp (f_{\btheta}(\bx)[y])~.
\end{equation}

\paragraph{The CEBM approach}~\cite{du2019implicit,du2020improved} learns a conditional energy function $E_{\btheta}(\bx | c)$.
Although CEBMs are mainly designed for image generation, they have also shown solid classification performance~\cite{du2019implicit} by using the class-conditioned energy of an image to predict its label:
\abovedisplayskip=3pt
\belowdisplayskip=3pt
\begin{equation}
\label{eq:experiments_cebm}
    y^\star ~=~ \argmin_{y} E_{\btheta}(\bx | y) ~.
\end{equation}
The generative objective of CEBMs lacks a clear stopping criterion.
During training, we monitor the model's FID score~\cite{heusel2017gans} and halt the optimization when the FID becomes stable and reaches a value indicating that generated images are visually similar to the training data.

\paragraph{Calibration of uncertainty.}
We also propose to look at the models' calibration since this was previously shown to be improved by energy-based training~\cite{grathwohl2019your,he2021joint}.
A model is considered well-calibrated if its confidence (\eg top softmax score) is higher for correct predictions than incorrect ones.
We use the standard \emph{expected~calibration~error}~(ECE) metric~\cite{guo2017calibration}. To compute the ECE, one splits the predictions into $M$ bins according to their confidence, and measures the weighted average of the difference between accuracy and confidence of each bin as follows:
\begin{align}
\label{eq:experiments_ece_1}
    \text{acc}(B_{m}) &~=~ {(1\,/\, |B_{m}|)} ~~\Sigma_{i \in B_m} 1(\hat{y}_i = y_i)\\
    \text{conf}(B_{m}) &~=~ {(1\,/\, |B_{m}|)} ~~\Sigma_{i \in B_m} \hat{p}_i\nonumber\\
    \text{ECE} &~=~ \Sigma_{m=1}^M \, (|B_{m}|/n) ~~ \big|\, \text{acc}(B_m) - \text{conf}(B_m) \,\big|\nonumber
\end{align}
where $B_m$ is the set of indices in bin $m$, $n$ is the total number of instances, $y_i$ and $\hat{y}_i$ are the predicted and ground truth labels for instance $i$, and $\hat{p}_i$ is the individual confidence (\eg softmax score) for instance $i$.
An ECE of 0 indicates perfect calibration \ie $p_i\!=\!1$ for correct predictions and 0 for incorrect ones.

\paragraph{Results of supervised energy-based training.} 
The baseline model (\tabref{tab:experiments_ebm_res}, second row) unsurprisingly overfits the object/color combinations of the training data and is therefore unable to generalize to novel combinations at test time.
The model obtains an accuracy of $100\%$ on questions about spheres (same combinations in training and test) but $0\%$ on cubes and cylinders (OOD combinations).
The calibration is also poor (high ECE of $71.54$\%) indicating that the model is confident about its wrong predictions.

\begin{figure}[h!]
	\centering
	\includegraphics[width=0.2\linewidth]{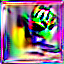}
	\includegraphics[width=0.2\linewidth]{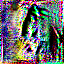}
	\includegraphics[width=0.2\linewidth]{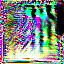}
	\includegraphics[width=0.2\linewidth]{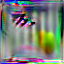}
	\\
	\includegraphics[width=0.2\linewidth]{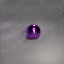}
	\includegraphics[width=0.2\linewidth]{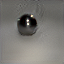}
	\includegraphics[width=0.2\linewidth]{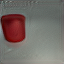}
	\includegraphics[width=0.2\linewidth]{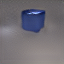}
  \caption{Generated images of high (top) and low energy (bottom).}
  \label{fig:experiments_gen}
\end{figure}

\paragraph{The standard JEM} (\tabref{tab:experiments_ebm_res}, third row) shows poor accuracy and calibration similar to the baseline.
Careful inspection revealed that the classification objective quickly dominates the energy learning.
In other words, the model could not simultaneously predict correct answers and be capable of generating new images.
The classification loss would typically converge much faster and, to minimize the energy loss, the model would start generating non-realistic images that always give a high energy.
The energy learning objective is thus ineffective in this regime.
We mitigate the dominance of the classification objective by down-scaling the corresponding term by a constant factor (0.1 proved effective).
The resulting model then learns to produce naturally-looking images (see \figref{fig:experiments_gen}, bottom row).
Nevertheless, at some point during training, we always observed that the model would start generating noisy images that quickly diverge from the training data (see \figref{fig:experiments_gen}, top row).
We solve this issue with early stopping, which we trigger when the energies of real and generated images diverge past a threshold: $|E_{\btheta}(\bx')-E_{\btheta}(\bx)|>0.8$.

\paragraph{The modified JEM} (\tabref{tab:experiments_ebm_res}, fourth row) gives a higher overall accuracy ($49.10\%$) as well as a better calibration (ECE of $36.55$\%).
Contrary to the baseline, this model can therefore correctly handle some of the OOD data, although it still often confuses cubes and cylinders with relatively high confidence.

\paragraph{The CEBM} seems to surpass the baseline, but its results are in fact not much better than random predictions.
Indeed, even though the CEBM does not overfit to the spurious correlations in the training data like the baseline, but it also fails to correctly answer questions about spheres (the ID test data), which were handled perfectly by the baseline and the JEM.
Moreover, we observed a surprising lack of correlation between the CEBM's FID score (\ie its generative performance, which we use for early stopping) and its VQA accuracy (\ie its discriminative performance).
In fact, the answers predicted by the model fluctuate wildly across epochs.
The low ECE score of the CEBM ($9.24$\%) is simply due a low confidence for all (mostly wrong) predictions. Thus it does not correspond to a desirable behavior.

\section{Conclusions}
\label{sec:conclusions}

The first part of this paper explored the applicability of self-supervised learning to pretrain visual representations for VQA.
We experimented with two instantiations of contrastive learning (CL) and energy-based models (EBMs).
With both, we could learn representations from unlabeled images that proved effective in a toy VQA setting.
After pretraining, the VQA models could be trained with little annotated data and handle OOD instances to some extent.
However, we found EBMs to be difficult to put into practice because of unstable training and variability in their results.
On the opposite, CL proved much more practical. In our setting, representations learned with CL on little data could compete with those from a supervised, ImageNet-pretrained model.

Overall, self-supervised pretraining is an increasingly attractive alternative to traditional transfer learning from a supervised model.
This option is particularly relevant when dealing with particular visual domains \eg for medical VQA~\cite{lau2018dataset,he2020pathvqa}.

In the second part of this paper, we investigated various purported benefits of EBMs (OOD detection, supervised energy-based training, calibration). These experiments were unfortunately largely negative.
This study is limited by the implementation choices that were necessarily made, and by our simplistic evaluation setting --~admittedly a far cry from open-domain VQA.
Still, our experience suggests that the state of the art on EBMs at the time of this study (mid 2021) is not ready for complex tasks like VQA. Other choices and future developments may of course affect this recommendation.

{\small
\bibliographystyle{ieee_fullname}
\bibliography{bibliography}}

\clearpage
\onecolumn
\appendix

\section*{Appendices}
\vspace{12pt}

\section{Derivation of \eqnref{eq:method_ebm}}
\label{appendix:ebm}
\noindent
The log of the EBM objective is:
\setlength{\abovedisplayskip}{15pt}
\setlength{\belowdisplayskip}{15pt}
\begin{equation}
\label{eq:log}
    \log (p_{\btheta}(\bx)) = E_{\btheta}(f_{\btheta}(\bx)))- \log(Z(\btheta)) \,,
\end{equation}
and the gradient of the normalizer is:
{\begin{align}
\label{eq:log_z}
    \frac{\partial}{\partial \btheta}\log\left(Z(\btheta)\right) &= \frac{\partial }{\partial \btheta} \log\left(\int \exp(E_{\btheta}(f_{\btheta}(\bx)))d\bx\right) \nonumber \\
    &= \frac{\partial }{\partial \btheta} \log\left(\int \exp(E_{\btheta}(f_{\btheta}(\bx)))d\bx\right) \nonumber \\
    &= \frac{\frac{\partial }{\partial \btheta} \left(\int \exp(E_{\btheta}(f_{\btheta}(\bx)))d\bx\right)}{\int \exp(E_{\btheta}(f_{\btheta}(\bx)))d\bx} \nonumber\\
    &= \frac{\int \frac{\partial }{\partial \btheta}\exp(E_{\btheta}(f_{\btheta}(\bx)))d\bx}{\int \exp(E_{\btheta}(f_{\btheta}(\bx)))d\bx} \nonumber\\
    &= \int \frac{\frac{\partial }{\partial \btheta}\exp(E_{\btheta}(f_{\btheta}(\bx)))}{\int \exp(E_{\btheta}(f_{\btheta}(\bx)))d\bx} d\bx \nonumber\\
    &= \int \frac{\frac{\partial }{\partial \btheta}E_{\btheta}(f_{\btheta}(\bx)) \exp(E_{\btheta}(f_{\btheta}(\bx)))}{\int \exp(E_{\btheta}(f_{\btheta}(\bx)))d\bx} d\bx \nonumber\\
    &= \int \frac{\partial }{\partial \btheta}E_{\btheta}(f_{\btheta}(\bx)) \frac{\exp(E_{\btheta}(f_{\btheta}(\bx)))}{\int \exp(E_{\btheta}(f_{\btheta}(\bx)))d\bx} d\bx \nonumber \\
    &= \int  p_{\btheta}(\bx) \frac{\partial }{\partial \btheta}E_{\btheta}(f_{\btheta}(\bx)) \nonumber \\
    &= \mathbb{E}_{\bx\sim p_{\btheta}(\bx)}\left[ \frac{\partial }{\partial \btheta}E_{\btheta}(f_{\btheta}(\bx)) \right]
    \,,
\end{align}}
Thus, the gradient of the log probability is:
{\small\begin{align}
\label{eq:method_ebm_appendix}
    \frac{\partial \log p_{\btheta}(\bx)}{\partial \btheta} &= \frac{\partial E_{\btheta}(f_{\btheta}(\bx)))}{\partial \btheta} -
    \mathbb{E}_{\bx'\sim p_{\btheta}(\bx')}\left[ \frac{\partial E_{\btheta}(f_{\btheta}(\bx'))}{\partial\btheta} \right]  \,,
\end{align}}
\end{document}

%% file: settings.tex
\newcommand{\figref}[1]{Figure~\ref{#1}}
\newcommand{\eqnref}[1]{Equation~\ref{#1}}

\newcommand{\sectref}[1]{Section~\ref{#1}}

\newcommand{\tabref}[1]{Table~\ref{#1}}
\newcommand{\appref}[1]{Appendix~\ref{#1}}

\newcommand{\cardinality}[1]{{#1}}

\DeclareMathOperator*{\argmin}{arg\,min}
\DeclareMathOperator*{\softmax}{softmax}

\newcommand{\ba}{\boldsymbol{a}}

\newcommand{\bp}{{\boldsymbol{p}}}
\newcommand{\bq}{{\boldsymbol{q}}}

\newcommand{\bv}{{\boldsymbol{v}}}

\newcommand{\bx}{{\boldsymbol{x}}}
\newcommand{\by}{{\boldsymbol{y}}}

\newcommand{\btheta}{{\boldsymbol{\theta}}}

%% file: background.tex
\subsection{A common self-learning objective}
\label{subsec:general}

The self-supervised learning objective of the interest in this report can be described as:
\begin{align}
\label{eq:general}
    \mathcal{L}_{\btheta}(\bx) ~&=~ -\!\mathbb{E}_{\bx\sim\mathcal{D}} \big[\log\left( \tilde{p}_{\btheta}(\bx) \right)\big]
    ~,~~ \text{with}\nonumber \\
    \tilde{p}_{\btheta}(\bx) ~&=~ \frac{\exp{\big(\text{score}(f_{\btheta}(\bx)))\big)}}{Z(\btheta)}
     ~,
\end{align}
where $\bx \in \mathbb{R}^\cardinality{D}$ is an input (unlabeled image), $f_{\btheta}: \mathbb{R}^\cardinality{D} \rightarrow \mathbb{R}^\cardinality{d}$ is the neural network of parameters $\btheta$ that learns a mapping from inputs to an embedding space, $Z(\btheta)$ is a normalizer and $\mathcal{D}$ is our dataset of unlabeled examples. Different choices of score function and normalizer give rise to different self-supervised learning algorithms. In particular, considering EBMs \cite{xiao2020vaebm} and the popular SimCLR~\cite{chen2020simple} method for CL:
\begin{itemize}

    \item The \textbf{score function} in EBMs is learnable. It is implemented as a separate neural network that maps an embedding to a scalar that represents an unnormalized density of the input.
    SimCLR uses a predetermined score function. It uses the similarity of the embeddings of different views (\ie semantic-preserving transformations) of the input.
    
    \item The \textbf{normalizer} in EBMs is designed to ensure $\int \tilde{p}(\bx)d\bx=1$. One approach to accomplish this is to estimate the normalizer using sampling, \ie learning to generate samples. 
    This is practically challenging and produces training instability.
    In SimCLR, rather than generating samples, augmented views of the training samples are used to compute $Z(\btheta)$. The estimation of $Z(\btheta)$ is simpler in SimCLR but the choice of the samples use is important.
    
    
\end{itemize}

\subsection{Energy-based models}
\label{subsec:method_ebm}
Energy-based models (EBMs) are a form of generative model that relies on an energy function to capture dependencies between the random variables to be modeled. The energy function maps each configuration of variables to a scalar energy value. It is optimized such that observed values (\ie the training examples) have a low energy \ie a high probability.
The probability density for an input $\bx \in \mathbb{R}^\cardinality{D}$ is represented as 
\abovedisplayskip=8pt
\belowdisplayskip=5pt
\begin{equation}
\label{eq:method_ebm_pdf}
    \text{score}(f_{\btheta}(\bx))) = E_{\btheta}(f_{\btheta}(\bx))) ~,
\end{equation}
where $E_{\btheta}: \mathbb{R}^\cardinality{d} \rightarrow \mathbb{R}$ is the energy function (with abuse of notation, we denote all parameters with $\btheta$), and $Z(\btheta)\!\!\!~=~\!\!\!\!\int \exp{(E_{\btheta}(f_{\btheta}(\bx)))} \,d\bx$ is the partition function. With this definition of $Z$, we have ${p}_{\btheta}(\bx)=\tilde{p}_{\btheta}(\bx)={\exp{(E_{\btheta}(f_{\btheta}(\bx)))}}/{Z(\btheta)}$, a proper density function that represents the distribution of input data.
In this work, we implement the energy function with a CNN that takes an image as input and returns a scalar.

\begin{figure}[h!]
  \centering
  \includegraphics[width=1\linewidth]{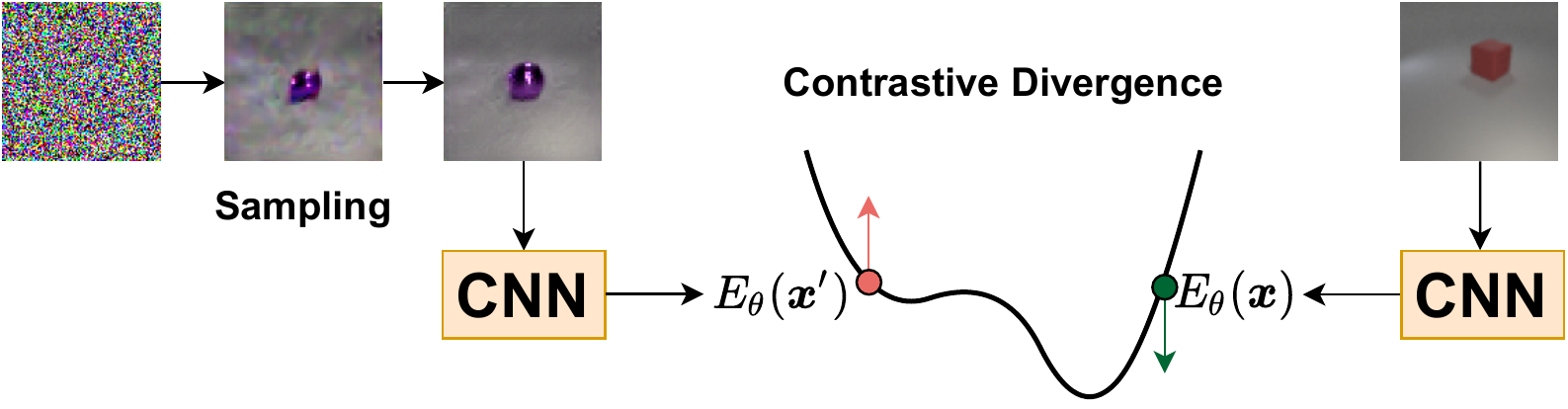}
  \caption{High-level illustration of the training of an EBM.} 
  \label{fig:method_ebm}
\end{figure}

The computation of the partition function $Z(\btheta)$ is usually intractable.
The standard maximum likelihood approach cannot therefore be applied directly to train the model.
Instead, we can use gradient-based optimization using the derivative of the log-likelihood of a sample $\bx$ (see \appref{appendix:ebm} for the full derivation):
{\small\begin{align}
\label{eq:method_ebm}
    \frac{\partial \log p_{\btheta}(\bx)}{\partial \btheta} &= \frac{\partial E_{\btheta}(f_{\btheta}(\bx)))}{\partial \btheta} -
    \mathbb{E}_{\bx'\sim p_{\btheta}(\bx')}\left[ \frac{\partial E_{\btheta}(f_{\btheta}(\bx'))}{\partial\btheta} \right]  \,,
\end{align}}%
where $\bx'$ is sampled from the model distribution.
Recent works~\cite{du2019implicit,nijkamp2019learning} have proposed to estimate this gradient with MCMC sampling (Markov Chain Monte Carlo). Using a number of $K$ samples, this gives:
\abovedisplayskip=8pt
\belowdisplayskip=6pt
{\small\begin{align}
\label{eq:method_ebm_log}
    \frac{\partial \log p_{\btheta}(\bx)}{\partial \btheta} &\approx
    \frac{\partial E_{\btheta}(f_{\btheta}(\bx)))}{\partial \btheta} - 
    \frac{1}{K}\sum_k\frac{\partial E_{\btheta}(f_{\btheta}(\bx_k')))}{\partial \btheta}\,,
\end{align}}
In this estimation, a single sample is typically obtained using Langevin dynamics~\cite{welling2011bayesian} from $p_{\btheta}$ as follows:
\abovedisplayskip=8pt
\belowdisplayskip=6pt
\begin{equation}    
\label{eq:method_ebm_mcmc}
    \bx'_{k,t} \,=\, \bx'_{k,t-1} \,+\, \frac{\lambda}{2} \frac{\partial E_{\btheta}(f_{\btheta}(\bx'_{k,t-1}))}{\partial \bx'_{k}} \,+\, \omega_{t}\,,
\end{equation}
with $\bx'_{k,t}$ denoting the $t$th iteration in the Markov chain for generating the $k$th instance, $\lambda$ being the step size, $\omega_{t} \sim \mathcal{N}(0, \lambda)$, and $\bx'_{k,0}$ initialized as uniform random noise.
The procedure defines a distribution $q_{\btheta}$ such that, if $t \!\!\rightarrow\! \infty$ and $\lambda \!\rightarrow\! 0$, then $q_{\btheta} \!\rightarrow\! p_{\btheta}$~\cite{welling2011bayesian}.
In practice, we differentiate only through the last step to reduce the computational cost as in~\cite{du2020improved}. The model is trained with a contrastive divergence objective~\cite{hinton2002training}. This minimizes the energy of the training data and maximizes that of the generated samples, as illustrated in \figref{fig:method_ebm}. The alternatives for training EBMs where the normalizer is approximated using a generator network is proposed in \cite{gade} that helps mitigate part of these practical limitations. 

\subsection{Contrastive learning}
\label{subsec:method_contr}

\begin{figure}[h!]
  \centering
  \includegraphics[width=1\linewidth]{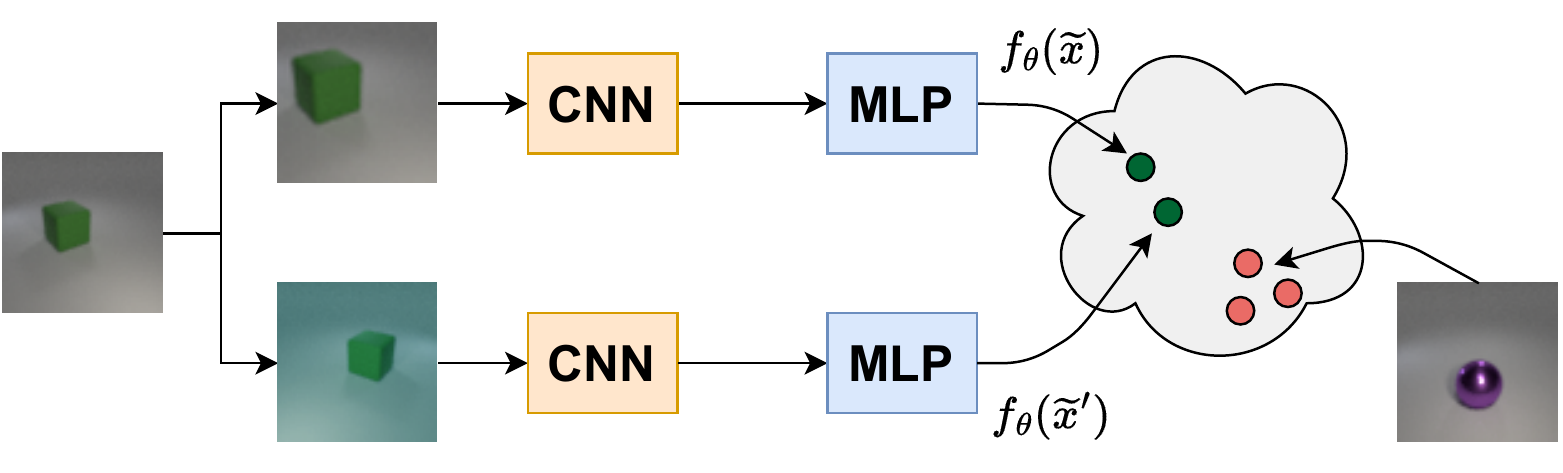}
  \caption{High-level illustration of contrastive learning.} 
  \label{fig:method_contrastive}
\end{figure}

Contrastive learning (CL) is another self-supervised technique whose core idea is to learn data representations such that similar samples are grouped together and dissimilar ones are pushed apart. In the absence of ground-truth annotations, all training examples are considered ``dissimilar'' from one another. ``Similar'' pairs are obtained by generating several views of each example with data augmentations (\ie hard-coded transformations that do not affect the semantic contents of the image).
The key components of CL are
\begin{enumerate*}[label=(\arabic*)]
    \item the data augmentations,
    \item the encoding function mapping pixels to the embedding space, and
    \item the contrastive loss that maximizes the similarity between encodings of similar pairs and minimizes that of dissimilar ones, as illustrated in \figref{fig:method_contrastive}.
\end{enumerate*}

On a high level, CL proceeds as follows.
Each training example $\bx$ is transformed with two sets of augmentations into a pair of views $\widetilde \bx$ and $\widetilde \bx'$.
They are passed through an encoding function such as a CNN to obtain representations. 
These form a \emph{positive} pair, while combinations of views from different examples form \emph{negative} pairs.
The training objective then maximizes the similarity of positive pairs while minimizing that of negative ones.
A common implementation of this objective is the normalized temperature-scaled cross-entropy loss over the cosine similarity of representations:
\abovedisplayskip=4pt
\belowdisplayskip=8pt
\begin{equation}
\label{eq:method_contr_sim}
    \text{score}(\widetilde\bx, \widetilde\bx') ~=~ \frac{f_{\btheta}(\widetilde\bx)^\top f_{\btheta}(\widetilde\bx')}{\tau\|f_{\btheta}(\widetilde\bx)\| \|f_{\btheta}(\widetilde\bx')\|} ~,
\end{equation}
where $\tau$ is a temperature hyperparameter. Further, 
\abovedisplayskip=4pt
\belowdisplayskip=0pt
\begin{align}
\label{eq:method_contr_loss}
     Z_{\text{CL}}(\btheta) &~=~ {\sum_{k}^{K} \exp(\text{score}(f_{\btheta}(\widetilde\bx), f_{\btheta}(\widetilde\bx_k)))} 
      \quad \bx\neq\bx_k~.\nonumber
\end{align}
Here, $\bx_k$ samples are usually chosen to be paired with the current mini-batch during SGD.
The overall loss is the sum  over all positive pairs in the current mini-batch.

%% file: Sections/tab_dataset.tex
\begin{table}[bth]
    \small
    \centering
    \renewcommand{\arraystretch}{1}
    \renewcommand\tabcolsep{7pt}
    \begin{tabular}{lccc}
          \toprule
          Split & Object/color & Nb. of & Nb. of\\
          ~ & combinations & images & questions\\
          \midrule
          Pretraining & $\mathcal{A}$ & 12,000 & --- \\
          Training & $\mathcal{B}$ & ~3,600 & 3,600 \\
          Validation & $\mathcal{B}\cup\mathcal{C}$ & ~~~800 & ~~800 \\
          Test & $\mathcal{C}$ & ~3,600 & 3,600 \\
          \bottomrule
    \end{tabular}
    \\\vspace{5pt}
    \textit{(a) Data splits.}
    \\\vspace{10pt}
    \renewcommand\tabcolsep{12pt}
    \begin{tabular}{clc}
              \toprule
              ~ & Object & Color\\
              \midrule
              \multirow{3}{1em}{$\mathcal{A}$} & Cylinder & Any.\\
              & Cube & Any.\\
              & Sphere & Any.\\
              \midrule
              \multirow{2}{1em}{$\mathcal{B}$} & Cube & Gray, blue, brown, yellow.\\
              & Sphere & Red, green, purple, cyan.\\
              \midrule
              \multirow{2}{1em}{$\mathcal{C}$} & Cube & Red, green, purple, cyan.\\
              & Sphere & Gray, blue, brown, yellow.\\
              \bottomrule
    \end{tabular}
    \\\vspace{5pt}
    \textit{(b) Sets of object/color combinations.}
    \\
    \caption{Summary of our diagnostic dataset. The pretraining, fine-tuning, and test splits contain different object/color combinations.}
    \label{tab:experiments_dataset}
\end{table}

%% file: Sections/fig_dataset.tex
\begin{figure}[htp]
  \centering
  \includegraphics{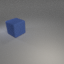}
  \includegraphics{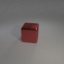}
  \includegraphics{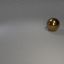}
  \includegraphics{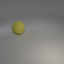}
  \includegraphics{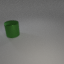}
  \includegraphics{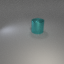}
  \caption{Example images from our diagnostic dataset.
  Questions are designed to query the object's shape. For example, for the first image: \textit{There is a blue object in the image; what is it?}, with the correct answer being \textit{cube}.
  } 
  \label{fig:experiments_data}
\end{figure}

%% file: Sections/tab_accuracy.tex
\begin{table*}[tb]
    \small
    \centering
    \renewcommand{\arraystretch}{1}
    \renewcommand\tabcolsep{4pt}
    \begin{tabular}{l@{\hskip25pt} c c@{\hskip12pt} c@{\hskip12pt} c@{\hskip12pt} c c}
          \toprule
          & Validation & ~ & \multicolumn{3}{c}{Test}\\
          \cmidrule{2-2}\cmidrule{4-6}
          & Overall && Overall & Cube only & Sphere only \\
          \midrule
          Baseline without visual pretraining & 56.96 \footnotesize$\pm\,$6.08 && 12.78 \footnotesize$\pm\,$9.75 & 0.04 \footnotesize$\pm\,$0.06 & 25.52 \footnotesize$\pm\,$19.57 \\
          Baseline with supervised ImageNet pretraining & 99.43 \footnotesize$\pm\,$0.33 && 99.14 \footnotesize$\pm\,$0.20 & 98.35 \footnotesize$\pm\,$0.33 & \textbf{99.92} \footnotesize$\pm\,$0.09 \\
          \midrule
          With self-supervised pretraining: EBM & 93.67 \footnotesize$\pm\,$3.18 && 86.74 \footnotesize$\pm\,$8.34 & 96.57 \footnotesize$\pm\,$1.99 & 76.91 \footnotesize$\pm\,$14.75\\
          With self-supervised pretraining: SimCLR$\star$ & \textbf{99.75} \footnotesize$\pm\,$0.33 && \textbf{99.26} \footnotesize$\pm\,$0.29 & \textbf{98.87} \footnotesize$\pm\,$0.27 & 99.65 \footnotesize$\pm\,$0.32 \\
          \bottomrule
    \end{tabular}
    \caption{Main results (average accuracy in \% $\pm$ one standard deviation over 3 random seeds).
    The baseline overfits to spurious correlations in the training data and performs poorly on the OOD validation and test sets.
    The EBM pretraining improves generalization, but SimCLR$\star$ performs significantly better and achieves near-perfect accuracy.}
    \label{tab:experiments_acc}
\end{table*}

%% file: Sections/fig_ood_ex.tex
\begin{figure}[h!]
    \centering
    \begin{subfigure}[t]{0.22\linewidth}
    	\centering
    	\includegraphics[width=.6\linewidth]{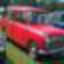}
    	\caption{\footnotesize CIFAR}
    \end{subfigure}
    \hspace{10pt}
    \begin{subfigure}[t]{0.22\linewidth}
    	\centering
    	\includegraphics[width=.6\linewidth]{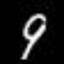}
    	\caption{\footnotesize MNIST}
    \end{subfigure}
    \hspace{10pt}
    \begin{subfigure}[t]{0.22\linewidth}
    	\centering
    	\includegraphics[width=.6\linewidth]{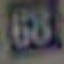}
    	\caption{\footnotesize SVHN}
    \end{subfigure}
    \\\vspace{4pt}
    \begin{subfigure}[b]{0.22\linewidth}
    	\centering
    	\includegraphics[width=.6\linewidth]{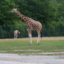}
    	\caption{\footnotesize VQA v2}
    \end{subfigure}
    \hspace{10pt}
    \begin{subfigure}[b]{0.22\linewidth}
    	\centering
    	\includegraphics[width=.6\linewidth]{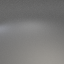}
    	\caption{\footnotesize Background}
    \end{subfigure}
    \hspace{10pt}
    \begin{subfigure}[b]{0.22\linewidth}
    	\centering
    	\includegraphics[width=.6\linewidth]{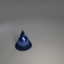}
    	\caption{\footnotesize Cones}
    \end{subfigure}
    \vspace{-4pt}\\
    \caption{Test examples used in our OOD detection experiment.} 
    \label{fig:experiments_ood_ex}
\end{figure}